\documentclass[10pt]{iopart}
\usepackage{graphicx}
\usepackage{hyperref}
\hypersetup{
    colorlinks=true,
    linkcolor=blue,
    filecolor=magenta,
    urlcolor=cyan,
    pdftitle={NaturalRoboticsContest},
    pdfpagemode=FullScreen,
    }

\begin{document}

\title[The Natural Robotics Contest]{The Natural Robotics Contest: Crowdsourced Biomimetic Design}

\author{Robert Siddall$^1$*, Raphael Zufferey$^2$, Sophie Armanini$^3$, Ketao Zhang$^4$, Sina Sareh$^5$ and Elisavetha Sergeev$^1$}

\address{$^1$University of Surrey, Guildford, UK. \\$^2$ École polytechnique fédérale de Lausanne (EPFL), Lausanne, Switzerland. \\$^3$ Technische Universität München, Munich Germany. \\$^4$  Queen Mary University of London, London, UK. \\$^5$ Royal College of Art, London, UK.}
\ead{*r.siddall@surrey.ac.uk}
\vspace{10pt}
\begin{indented}
\item[]October 19th 2022
\end{indented}

\begin{abstract}
Biomimetic and Bioinspired design is not only a potent resource for roboticists looking to develop robust engineering systems or understand the natural world. It is also a uniquely accessible entry point into science and technology. Every person on Earth constantly interacts with nature, and most people have an intuitive sense of animal and plant behavior, even without realizing it. The Natural Robotics Contest is novel piece of science communication that takes advantage of this intuition, and creates an opportunity for anyone with an interest in nature or robotics to submit their idea and have it turned into a real engineering system. In this paper we will discuss the competition's submissions, which show how the public thinks of nature as well as the problems people see as most pressing for engineers to solve. We will then show our design process from the winning submitted concept sketch through to functioning robot, to offer a case study in biomimetic robot design. The winning design is a robotic fish which uses gill structures to filter out microplastics. This was fabricated into an open source robot with a novel 3D printed gill design. By presenting the competition and the winning entry we hope to foster further interest in nature-inspired design, and increase the interplay between nature and engineering in the minds of readers.
\end{abstract}

%
\vspace{2pc}
\ioptwocol

\begin{figure*}[ht]
 	\centering
      \includegraphics[width=1\textwidth]{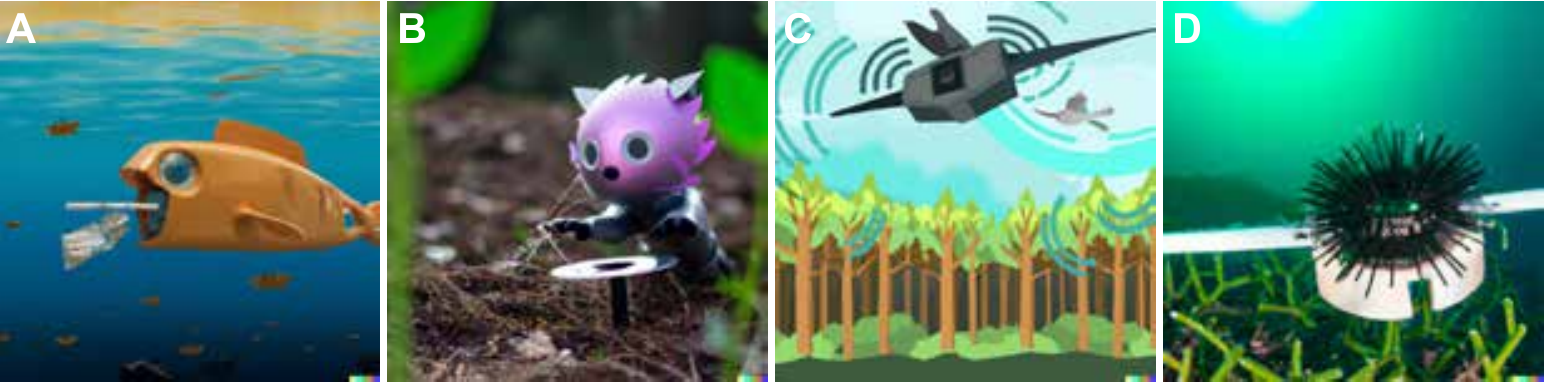}
      \caption{A selection of top-scoring ideas for bioinspired robots, rendered by Dall-E. A) A robotic fish ingests plastic waste from the ocean. B) A robotic squirrel plants milkweed seeds. C) A robotic bird patrols a forest to track deforestation. D) A robotic sea urchin cleans algae from coral and combats acidification with secretions.}
      \label{fig:DallE}
\end{figure*}

\section{Introduction}

Bioinspiration is the process of taking observations from naturally occurring systems and applying it to synthetic systems. Learning from nature is not new - for most of human history there was no obvious way to distinguish where the `natural' world stopped and the synthetic world began, and `bioinspiration' as a term would have been somewhat redundant. However, as many natural processes are pushed to the boundaries of society and the built environment occupies more of our reality, a need to resume the process of learning from nature has been felt within the scientific community. Moreover, technology now allows us to understand with far greater detail the processes and structures which underpin the dynamics of nature, and our improving understanding of evolution allows greater appreciation of the efficiencies and performance gains that have been wrought by eons of natural selection. Natural materials, movement and behaviour offer the means for technologists to find approaches that maximise the use of available resources, rather than relying on extractive means of increasing performance (e.g. the use of ever greater energy in producing and operating a system).

Robotics is a field which can draw particular benefit from the reservoir of evolved knowledge in the natural world, as it strives to build mechanical systems which face many of the same challenges as animals moving through the world. By looking at nature readers can find new modes of locomotion \cite{zufferey2022between}, understand the limits of performance and how to overcome them \cite{williams2009pitch}, and find small and almost costless means of improving performance \cite{fish2006passive}. Bioinspired design has also driven interest in the benefits of compliant structures \cite{siddall2021compliance} and plays an important role in the growing field of soft robotics \cite{kim2013soft}. Biomimetic robots that directly copy animals can even be used as means to better understand animals themselves, by functioning as physical models for biomechanics studies \cite{siddall2021tails}.

Nature is a fantastic entry-point for teaching \cite{full2021eyes}; almost everyone has an intuitive sense for animal behavior and locomotion from watching anything from movies to pets, pigeons, squirrels and other ubiquitous wildlife, even without realizing it. What is often needed is simply a way to think about what is already subconsciously known. By holding a bioinspired design competition, we provided a novel way for people to engage with creative design outside of a normal didactic environment, and documenting the winning design will provide a recent and tangible ‘case study’ to be used in teaching. And it was fun.

In this paper we will describe a public bioinspired design competition, `The Natural Robotics Contest'. We break down the types of ideas generated by participants (examples are given in figure \ref{fig:DallE}, rendered by Dall-E 2 \cite{ramesh2022hierarchical} for compactness, with original entries shown in \ref{sec:entries}). We will feature a selection of the best ideas selected by the competition judges (the authors of this manuscript), before presenting the process of turning the winning entry into a working prototype, and displaying the robot in-action.

\section{The Natural Robotics Contest}

The Natural Robotics Contest is novel piece of science communication, intended to be an opportunity for anyone with an interest in nature or robotics to have their idea turned into a real engineering system. The brief for the contest was simple - entrants needed to submit an idea for a robot, inspired by nature, that can do something to help the world (see \ref{sec:flyer}). The competition was marketed principally to high school and university students but entry was open to anyone interested. It has a deliberately low barrier to entry - only a simple sketch and description was asked for, so that it was accessible to entrants from all subjects and experience levels, and the website was explicit that the judging panel was looking for creativity and potential impact, not drawing ability. Over the two months that the competition was open to submissions, it received approximately 100 entries.

\subsection{Submitted Entries}

\begin{figure*}
 	\centering
      \includegraphics[width=1\textwidth]{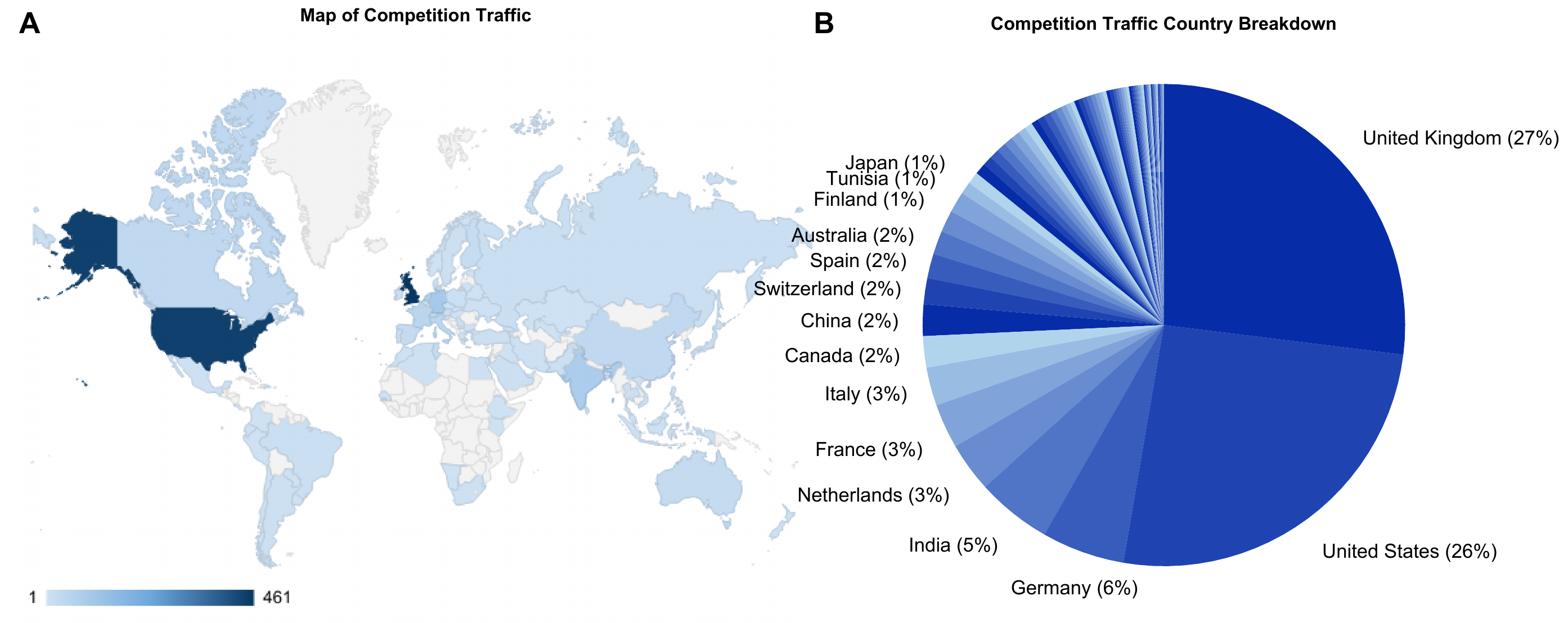}
      \caption{Contest web traffic, based on counts of unique users. A) Map of countries from which the contest webpage was accessed. B) Breakdown by country. The majority of traffic came from the UK and the USA.}
      \label{fig:AnalyticsData}
\end{figure*}

 In the interests of protecting privacy, personal data was not collected from the entrants beyond an email address for communication. However, website analytics provided an insight into the reach and interest in the competition from around the world (figure \ref{fig:AnalyticsData}). The majority of traffic came from the UK and USA, together accounting for around 50\% of the total. This is to be expected given the outlets the contest was promoted in, and the language of the website. The contest was also promoted in German, and as a consequence the third largest proportion of traffic came from Germany.

\begin{figure*}
 	\centering
      \includegraphics[width=1\textwidth]{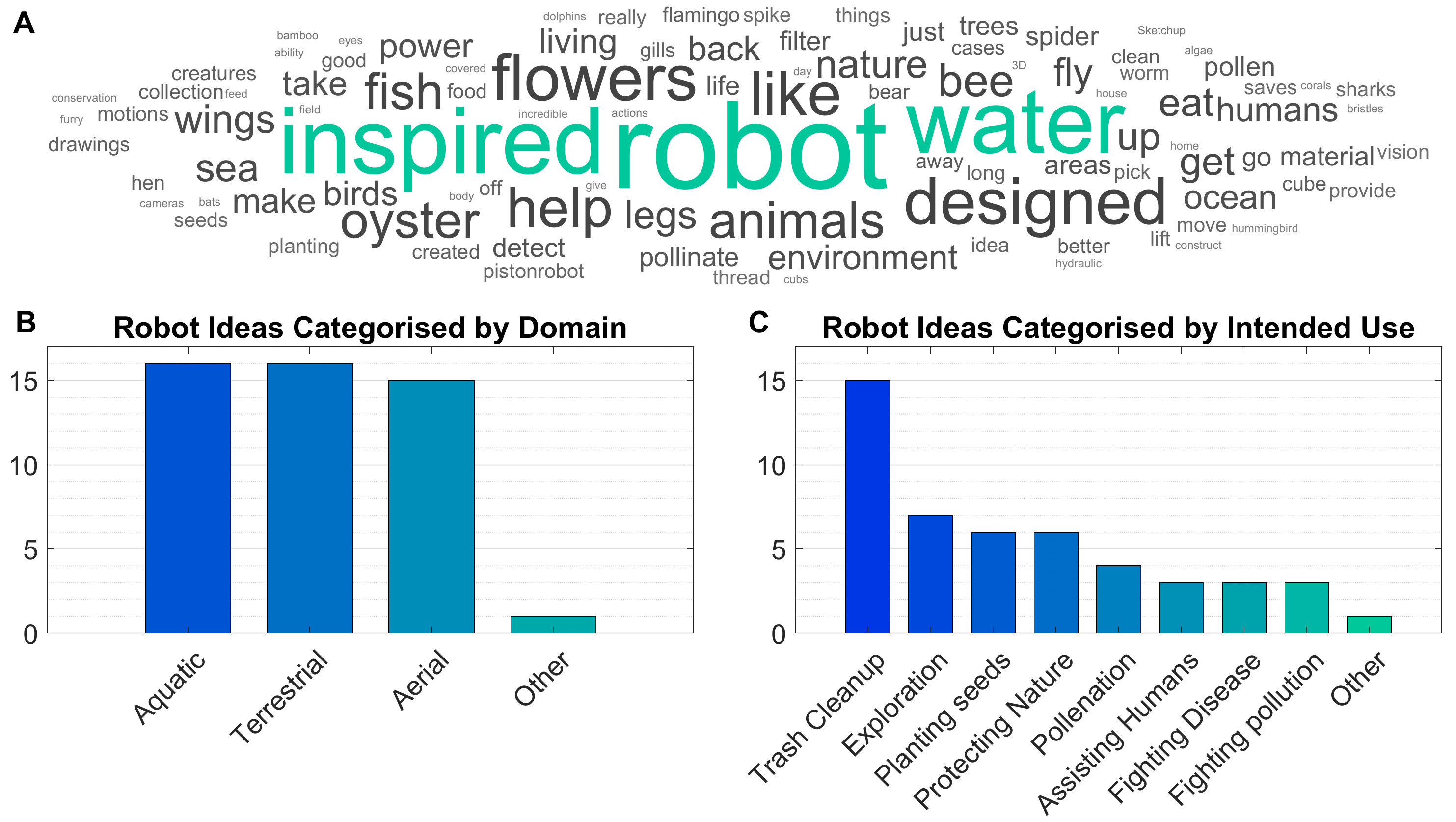}
      \caption{Summary of the types of idea submitted to the contest. A) Wordcloud of the descriptions submitted with all contest entries. B) Design ideas categorised by the domain they move in. C) Robot submissions categorised by intended use. While there was an even spread of ideas across air, land and water, robots designed to remove waste were by far the most common.}
      \label{fig:ContestData}
\end{figure*}

The contest received a wide variety of proposals taking inspiration from a diverse set of natural systems (figure \ref{fig:ContestData}). There was an even spread of robots across flight, swimming and terrestrial locomotion (figure \ref{fig:ContestData}B), but a pronounced preference among participants to design robots which could help to remove waste from the environment, in particular the ocean (figure \ref{fig:ContestData}C). The second most common type of design was a robot which provided some form of service to plant life, whether by pollinating, seed planting or otherwise monitoring and protecting forests and similar ecosystems.

While there is not enough space to cover every entry in this manuscript, some notable entries are discussed here, with the submitted drawings included in \ref{sec:entries}. `SkyRanger' by Teju Sankuratri (figure \ref{fig:skyranger}) proposes the use of biomimetic birds as a means to survey ecosystems and function as an early warning system for ecological harm. The development of bird-inspired robots is an active area of research \cite{zufferey2022ornithopters}, and while many design concepts have been proposed, practical application demonstrations remain scarce and there is significant further research effort needed. The proposed `Ersters' robot by Elizabeth Ivanova (figure \ref{fig:ersters}) is an excellent idea that identifies an important ecosystem service offered by oysters \cite{zu2013quantifying}, although the judges noted that in this instance it was not immediately obvious how a robot could improve upon the filtration already performed by the natural animals. The `Specialised Anti-Acidification Sea Urchin' by `The Robotineers' (figure \ref{fig:urchin}) was another well-researched idea for an ocean clean-up technology that was very popular with the judges. Sea Urchins are tenacious animals with profound effects on many ecosystems, and harnessing some of their adaptations to protect coral is an attractive idea. `Bumblebot' by Daniella Clifton (figure \ref{fig:bumblebot}), is one of several robotic pollinators proposed among the contest entries, and echoes the considerable interest in bee-inspired robots seen in the aerial robotics field, where one of the smallest and best-known robots is the Harvard Robobee \cite{ma2013controlled}.
The Hermit crab rover by `The Yak Collective' (figure \ref{fig:hermitcrab}) is a scavenger robot, which gathers scrap material from its surroundings to build itself a protective shell.  In a somewhat similar vein is the `Milkweed planting squirrel' by Sue Klefstad (figure \ref{fig:milkweed}), which seeks to emulate the seed burying behaviour of squirrels to plant milkweed, a plant which is essential to the lifecycle of many butterfly (\textit{Danainae}) species, including the monarch butterfly.
A number of submissions also focused exclusively on enabling new forms of locomotion to explore the robot's surroundings. An example is `Spider-Poppins' by Maier Fenster (figure \ref{fig:spiderpoppins}), which mimics the ballooning motion of spiders. Though scientific observations of spider ballooning date back to Charles Darwin \cite{gorham2013ballooning}, this is is a form of locomotion that has been explored in new detail by recent biological literature, and is only now beginning to be fully understood \cite{choballooning2020}.

\begin{figure*}
 	\centering
      \includegraphics[width=0.9\textwidth]{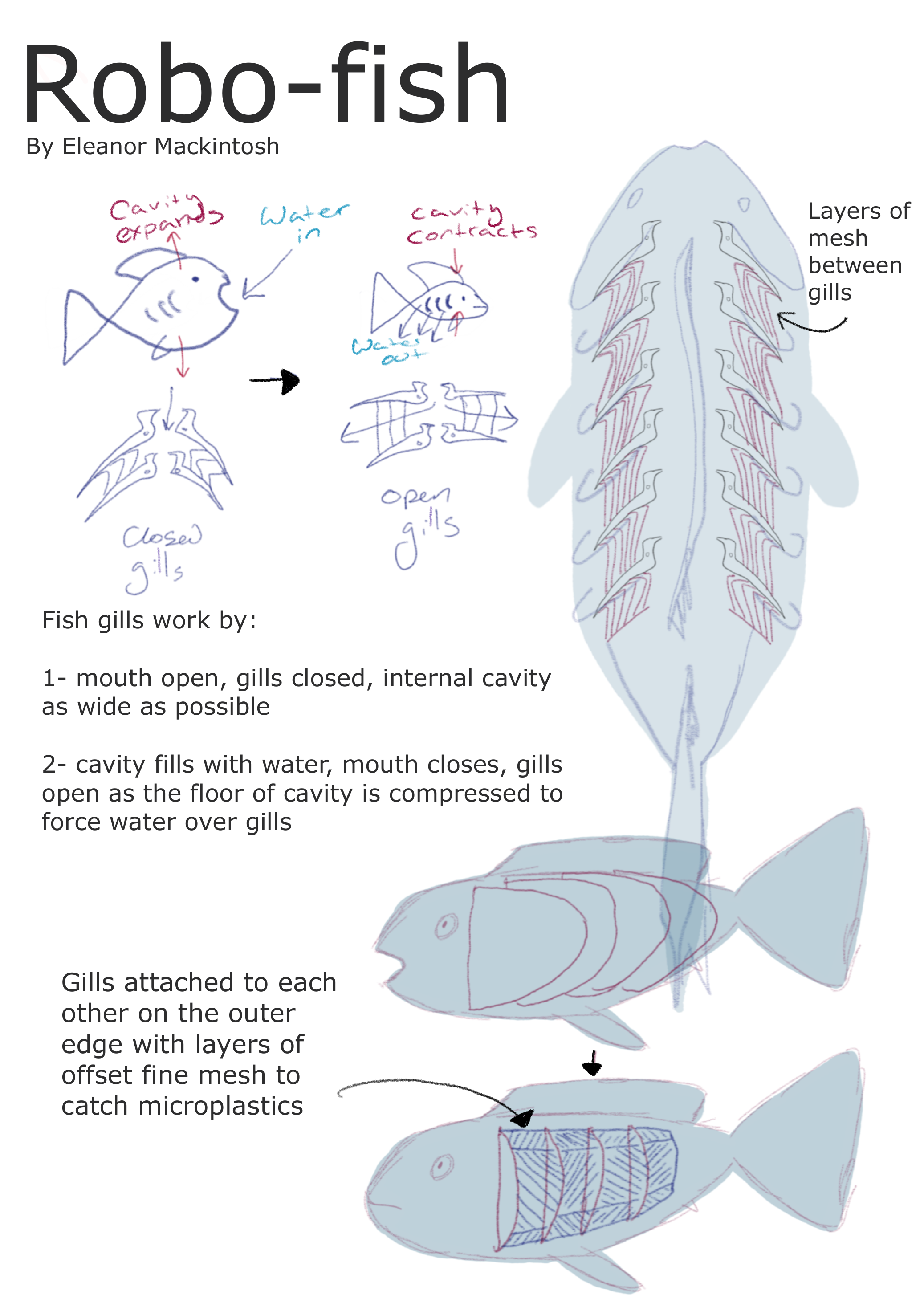}
      \caption{The winning contest entry, by Eleanor Mackintosh: A robotic fish which uses gills to filter and sample microplastic pollution in aquatic ecosystems.}
      \label{fig:Eleanor}
\end{figure*}

The submitted designs were all given a mark from 1-5 by each of the competition judges, and the design with the highest aggregate mark was selected as the winner. This year, three designs were tied for first place based on scores: Eleanor Mackintosh's `Robofish', Teju Sankuratri's `Sky Ranger' (figure \ref{fig:skyranger}) and the `Specialised Anti-Acidification Sea Urchin' by `The Robotineers' (figure \ref{fig:urchin}). The winner was selected from those three by the judges. Eleanor Mackintosh's idea for a microplastic filtering fish was ultimately chosen as the winner (figure \ref{fig:Eleanor}). This design was chosen not only for the detailed thought put into the design and application, but also because the robot's purpose as a tool for ocean clean-up represented the most commonly proposed use case across all competition entries (figure \ref{fig:ContestData}C).

\subsection{Analysis of a biomimetic system for removing micro-plastics}

Before developing the proposed idea into an engineering system, it was necessary to gain a better understanding of the problem it was trying to solve. Microplastic pollution is a growing global concern, and neither the geography nor the impact of the problem is well understood. Estimates of plastic concentrations vary widely, due to both a paucity of data and variability in sampling methods. Predictions by the World Economic Forum show that plastic could exceed fish by weight by 2050 \cite{agenda2016new}.

Removing extant ocean microplastic through robotic filtration is unlikely to be successful. There is simply no reliable way of distinguishing organic matter that is vital to the ecosystem such as plankton and `marine snow' from synthetic pollutants, and it is difficult to imagine how a cleanup could avoid directly harming marine life in the process.

Moreover, the scale of removal necessary is likely beyond the reach of current technology. If we take the example of one of the largest filter feeding marine organisms, the basking shark, we can get a sense of the problem. Basking sharks filter around 30kg of particulate from the water each day by filtering around 800m$^3$ of water per hour \cite{fossi2014large}. Filtering all ocean water would take 100 billion shark-years, and even if a all  of the 30kg of particulate matter were plastic (in actuality, microplastics are found at concentrations of a few particles per liter \cite{cressey2016plastic}), it would take 1 million basking sharks to filter out the 10 million tonnes \cite{jambeck2015plastic} of plastic entering the ocean each year.

Even systems engineered to remove plastic at scale struggle. Last year, in collaboration with researchers in the United Kingdom and Germany, Hohn \textit{et al.}\cite{hohn2020long} published an analysis of what it would take for the Ocean Cleanup to collect only the floating plastic in the largest five gyres. Hohn \textit{et al.} took the current amount of plastic in the ocean, added annual inputs, and compared it with how much plastic the Ocean Cleanup’s successful pilot collected. To clean up a fraction of one percent of the total, the Ocean Cleanup would have to run nonstop until 2150. Even when Hohn \textit{et al.} artificially increased the fleet to 200 booms, the project still only recovered five percent of the floating plastic \cite{hohn2020long}. However, while immediate removal of ocean plastic is not feasible, targeted removal efforts do have a significant effect. The aforementioned Ocean cleanup is currently deploying 'interceptors' to the world's most polluted rivers \cite{williams2020rid}, to prevent plastic from reaching the ocean.

So this is not to say that the problem is intractable, nor that there is no role for technology - the opposite is true. There is an immediate need for better data on microplastics, as the location of the vast majority of the plastic waste that has been dumped into aquatic ecosystems is unknown, and robots could play a leading role in this task. Targeted cleanups are effective mitigations \cite{conservancy201630th}, but need better data in order to be focused effectively and maximise resource use. This is especially true of freshwater ecosystems, which account for only ~4\% of published research on plastic waste \cite{royalsocmicroplastics2019}. A microplastic-filtering robot, particularly one which could access areas of the water inaccessible to humans, would be very useful as a data collection tool. This was identified by the competition winner in her entry.

\begin{figure*}
 	\centering
      \includegraphics[width=1\textwidth]{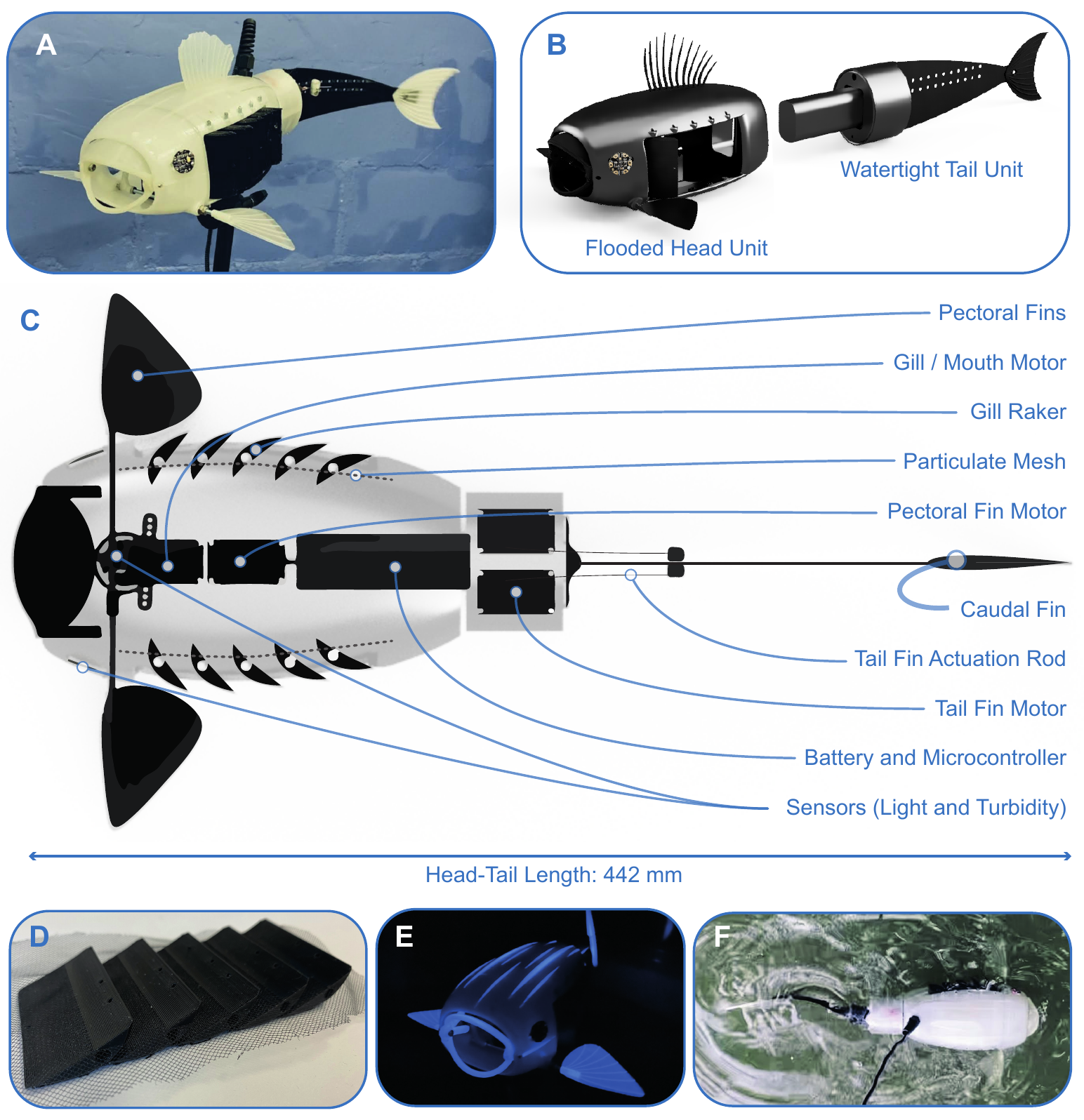}
      \caption{The fabricated contest winner, a microplastic-collecting robotic fish. A) Image of the completed fish. B) CAD render of the fish design, showing the modular design, with a separate, watertight unit for propulsion, power and control, and a swappable head unit for different missions. C) Diagram of the robot, showing the internal mechanics, including location of actuation motors, and the position of gill rakers and mesh for filtering. D) Image of the gill rakers, with mesh integrated directly into the 3D printing process. E) Image of the fish showing phosphorescent accents. F) Image of the fish swimming in a lake.}
      \label{fig:FishDiagram}
\end{figure*}

\section{Creating the Winning Design}

The winning design for a pollution-filtering robot with fish-like gills, was developed into a 442 mm long robotic fish (figure \ref{fig:FishDiagram}A-B), which moves by body-caudal fin undulation, with a carangiform propulsion mode. The winner of the contest was communicated to participants via an online video \footnote{\href{https://youtu.be/ld15OYvvgfk}{youtu.be/ld15OYvvgfk}}, which shows the robot in action. The robot has a large head cavity with an openable mouth and sets of gills that contain a 2 mm nylon mesh. The robot is remotely controlled, although it has been equipped with the sensors necessary for basic autonomy in future iterations.

Undulation of the tail is driven by motorised pushrods (made from 0.5 mm diameter music wire) connected to the base of the tail, a design which was used to good effect by \cite{van2022openfish}. Unlike \cite{van2022openfish}, the left and right pushrods are driven by separate motors (figure \ref{fig:FishDiagram}C), which both allows for steering via changing the relative amplitude of the left and right motors, and increased power while still using a popular and affordable smart servo model (XL330-M288). Finally using separate motors allows for a future iteration of the fish robot to make use of antagonistic co-contraction of the two swim motors, which has benefits to swimming efficiency \cite{lin2021modeling}. The robot is designed to be neutrally buoyant and uses actuated pectoral fins to control pitch and depth.

It was decided that the robot should only use affordable off-the-shelf components and manufacturing techniques, so that the design is accessible to all. As such, the robot is entirely 3D printed with a low-cost fused deposition manufacturing (FDM) printer (Prusa Mini+, 0.4 mm nozzle), with the control electronics, battery and propulsion motors contained in a sealed `tail' unit, onto which the `head' of the robot is attached via a snap-fit joint (figure \ref{fig:FishDiagram}B). This modular design was chosen so that the head could be readily changed to meet different gill arrangements in the future.

\begin{figure*}
 	\centering
      \includegraphics[width=1\textwidth]{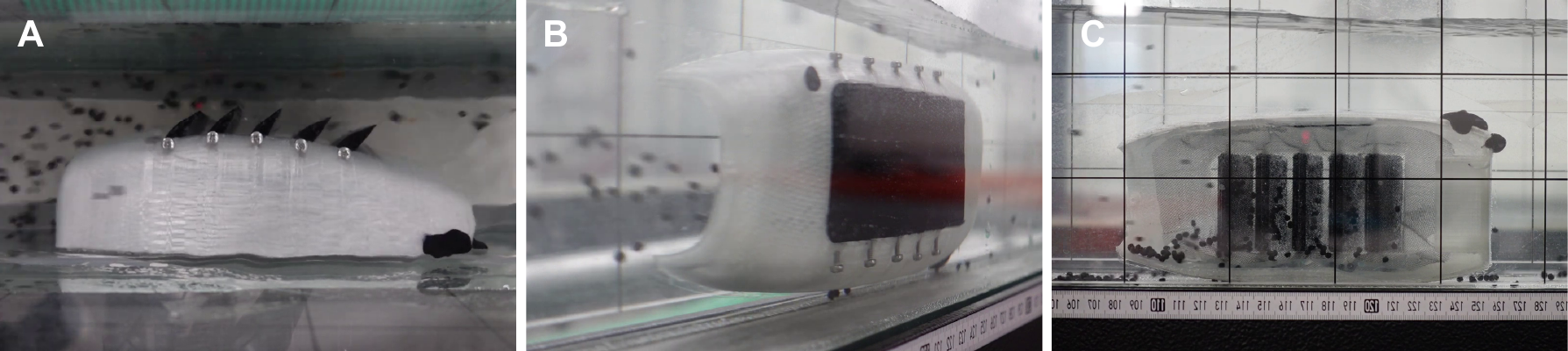}
      \caption{Demonstration of the fish's gills, used for filtering plastic particles. The robot's head was bisected along the sagittal plane and placed in a water flume. In all images, water flow direction is from left to right. A) Top view, showing gill angle (all gills are identically sized. B) Front view, showing mouth. C) internal view, showing particles trapped against gills by vortices shed from internal edges of the gill plates.}
      \label{fig:Gills}
\end{figure*}

The gill structures in ram-filter feeding marine animals typically have structures which obstruct the internal water flow. This creates trapped vortices behind each gill raker which aid the collection of particulate matter with less impediment to the flow of water through the gill structure \cite{sanderson2016fish}. To achieve this in a robot, we 3D printed an array of gill plates, and created an interstitial mesh by pausing the print and inserting nylon mesh between layers (figure \ref{fig:FishDiagram}D). Each gill rotates around a rod, and the gill array is opened and closed via a pushrod connected to the leading gill. A gill array was bisected along the sagittal plane and affixed to a transparent sheet (figure \ref{fig:Gills}) for testing in an 86 mm wide water flume (HM 160, GUNT Gerätebau GmbH). A 0.8 l/s flow rate was used, giving a mean water velocity of 8 cm/s at the mouth of the fish. When opened, the gill array passively collects incoming particles (figure \ref{fig:Gills}) and arrests their locomotion at the mesh, where they eventually drop to the stagnant area of flow at the base of the robot without accruing and obstructing the gills.

Neutral buoyancy is achieved by modifying the infill density of all 3D printed parts, such that the net weight of the part is as close to equal to its displaced volume as can be achieved without compromising watertightness. The majority of the robot is printed in ABS plastic which is dipped in acetone to seal the micropores resulting from the FDM printing process, which would otherwise cause leaks. For aesthetic reasons, selected components and additional accents are printed in phosphorescent PLA (figure \ref{fig:FishDiagram}E). The two motors which are outside the watertight tail section were disassembled and waterproofed by covering circuitry with an acrylic conformal coating (Electrolube Acrylic Protective Lacquer) and filling the entire servo cavity with non-conductive grease (Liqui Moly 3140). This method performed well and after an initial gasket failure that was repared, no ingress of liquid was observed in the tests performed, which included throwing the fish into the water from the shore around 10 times.

The robot uses four Dynamixel XL330 smart servos for actuation, with two XL330-M288 motors powering the caudal fin, and two XL330-M077 motors (with lower torque and higher speed than the M288 model) controlling the gills and pectoral fins (figure \ref{fig:FishDiagram}C). The robot is controlled over WiFi using a remote (Xbox One controller), via an Arduino Nano33 IoT microcontroller, which contains an LSM6DS3 inertial measurement unit. A turbidity sensor is placed inside the mouth of the fish to sense particulate concentration, and an exterior light/colour sensor (TCS34725) provides basic navigation cues. The robot is powered by a 5000mAh battery (Auskang USB-C power pack). The USB serial port of the microncontroller is connected to a wire, to allow easy reprogramming of the fish during testing (figure \ref{fig:FishDiagram}F) without opening watertight compartments. This will be removed in future versions of the robot.

The robot was testing in an outdoor lake in Guildford (UK), and demonstrated effective swimming and steering on the water surface. The typical swimming speed was 5 cm/s at a tailbeat frequency of 2 Hz, with the speed limited by the high drag of the robot's filtering head and the relatively small propulsion motors (3 W of propulsion power versus 10W in \cite{van2022openfish}). However speed could be improved with a better optimisation of the caudal fin and better matching of the robot's tail stiffness to the inertia of its head in future iterations. The robot was able to submerge, but a tendency of the mouth cavity to trap air was an issue. A future iteration would benefit from the provision of buoyancy control \cite{zufferey2022multirotor}. Fortunately, there is ample space in the head cavity for this (figure \ref{fig:FishDiagram}E)

\subsection{Future Work}

Currently, the robot is able to ingest and retain particulates, but has no means of analysing them directly. To be an effective tool for ocean sampling, this would need to be automated. As the tools to analyse microplastics (e.g. Fourier-transform infrared spectroscopy equipment) require rigid, calibrated optics, and do not currently minaturise well. The authors intend to develop a larger floating docking station, that could pump out collected material into a sampling chamber and clean the interior of the robot for a new sampling mission. Collected material could then be analysed while further samples are collected. This base station could also function as a charging point for the robot, as well as a repeater for wireless communications, ameliorating the difficulty of signal transmission through water.

\section{Conclusion}

The Natural Robotics Contest has collected ideas from around the world, and shows not only the desire among the public to improve nature with technology, but also a thoughtful approach to looking at nature among participants, with many innovate ideas on display. The winning robot has realised the design features proposed by its originator, and now offers a promising new application for biomimetic underwater robots, that will be developed further in future. This is the first iteration of the Natural Robotics Contest, and the authors plan to repeat the contest in coming years, with future version of the contest featuring more detailed design challenges that represent the most pressing needs of the day. By building a library of bioinspired design year on year, the contest will become a resource for those who wish to harness nature to improve the world.

\section{Contribution Statement}

The contest and all associated graphic/web design was created by RS. RS, KZ, SS, SA and RZ judged the contest. The winning entry was designed and fabricated by RS, LS, and RZ. RS prepared the paper, with all authors contributing to the final version.

\section{Data Availability}

More information on the competition is available on its website:  \href{https://www.naturalroboticscontest.com}{www.naturalroboticscontest.com}, and the CAD design for the presented robot is available for download: \href{https://grabcad.com/library/robotic-fish-5 }{grabcad.com/library/robotic-fish-5}. The winner of the contest was announced via an online video, which can be viewed here: \href{https://youtu.be/ld15OYvvgfk}{youtu.be/ld15OYvvgfk}. Any other information can be made available upon request.

\section{Acknowledgments}

The authors would like to thank all the entrants to the competition for their thoughtful and creative submissions. Thanks also go to Fabian Franke and the COMLAB4 team for helping to dream up the contest.  This project was funded by the Alexander von Humboldt Foundation, the International Journalist's Programmes, and the
University of Surrey's Teaching Innovation Fund.

\clearpage

\onecolumn
\appendix

\section{Selected Competition Entries} \label{sec:entries}

In recognition of the effort that went into producing designs by the contest's participants, we have included several notable entries as an appendix to this paper. While it was not possible to include every submission, the judges would like to note that almost every entry considered had merit, and it was ultimately a difficult decision.

\begin{figure*}[h]
 	\centering
      \includegraphics[width=1\textwidth]{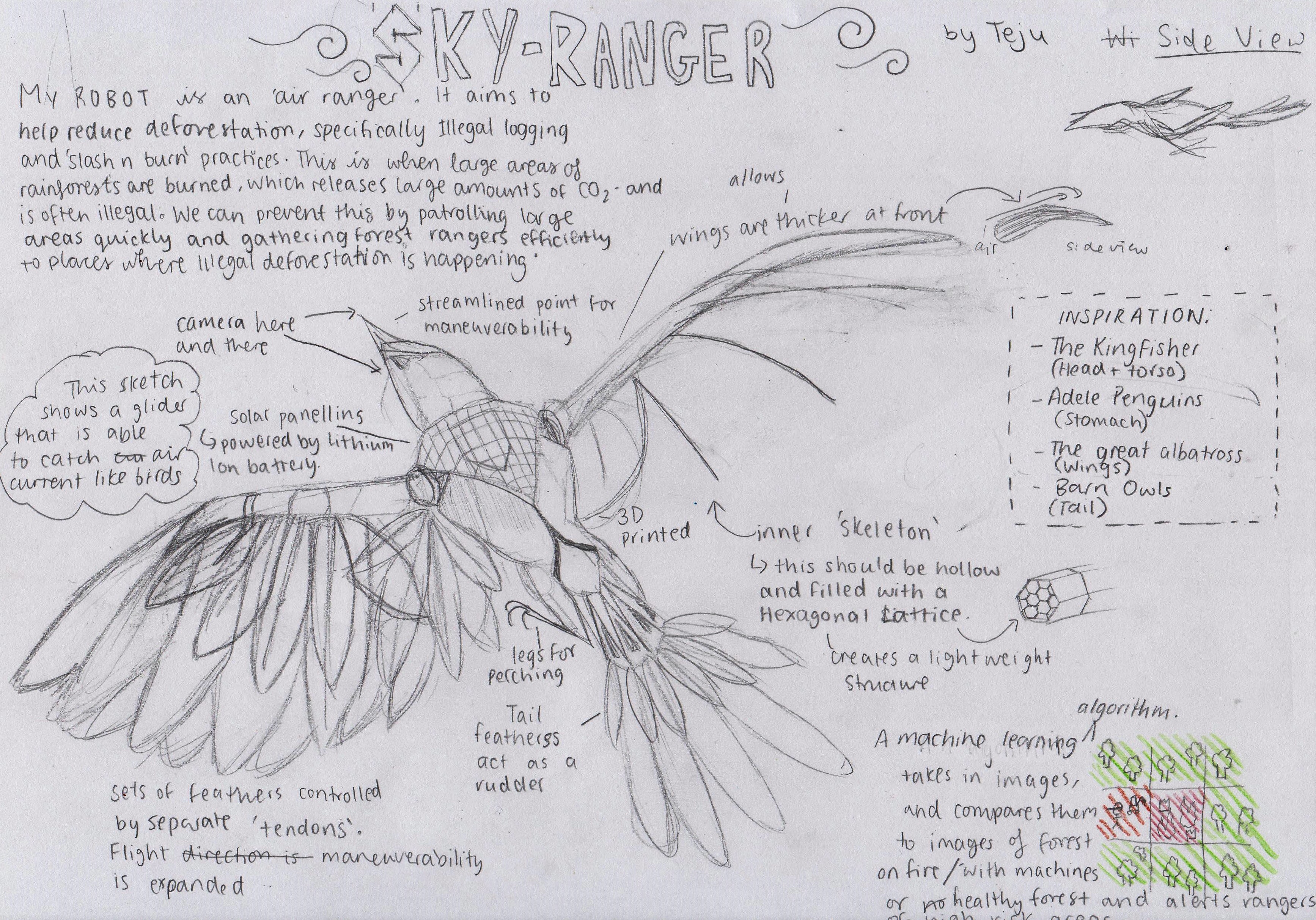}
      \caption{'Sky Ranger', a submission by Teju Sankuratri. The proposed robot is bird inspired, and intended to track and combat deforestation.}
      \label{fig:skyranger}
\end{figure*}

\begin{figure*}[h]
 	\centering
      \includegraphics[width=1\textwidth]{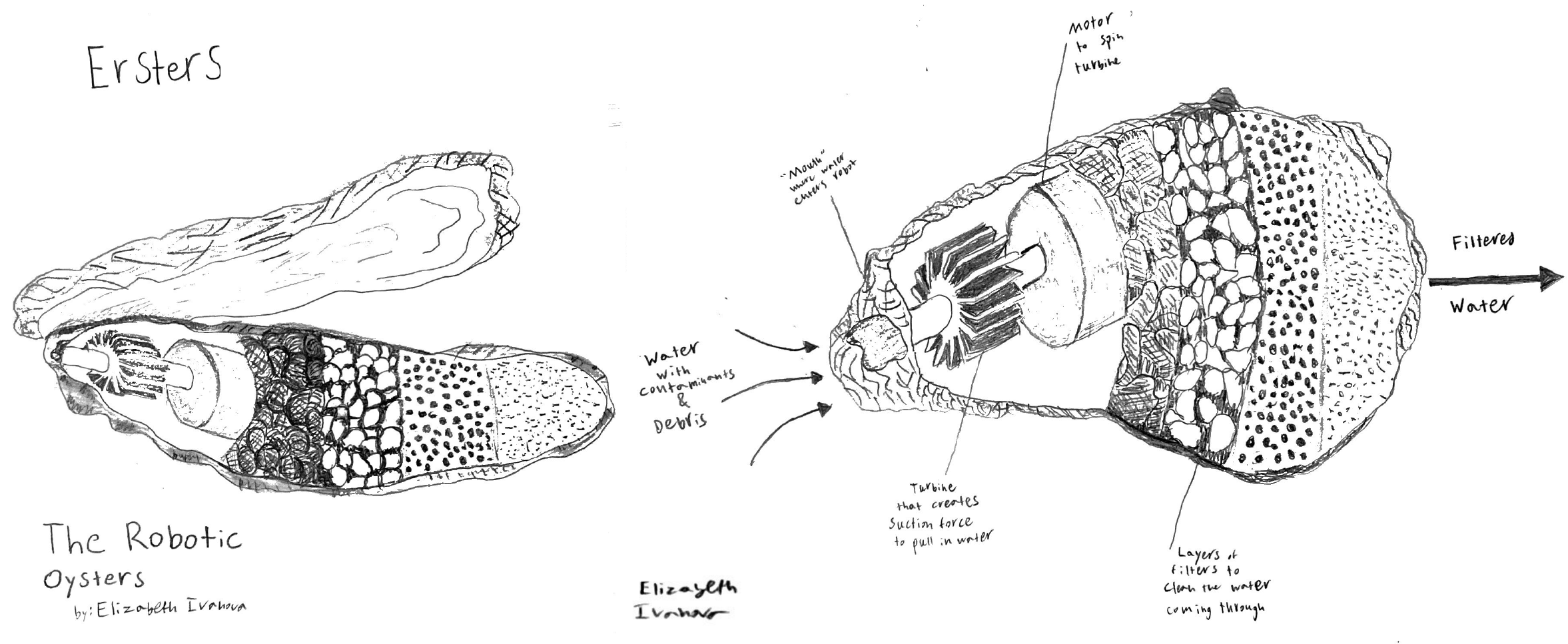}
      \caption{A robotic oyster used to filter/clean water, by Elizabeth Ivanova.}
      \label{fig:ersters}
\end{figure*}

\begin{figure*}[h]
 	\centering
      \includegraphics[width=1\textwidth]{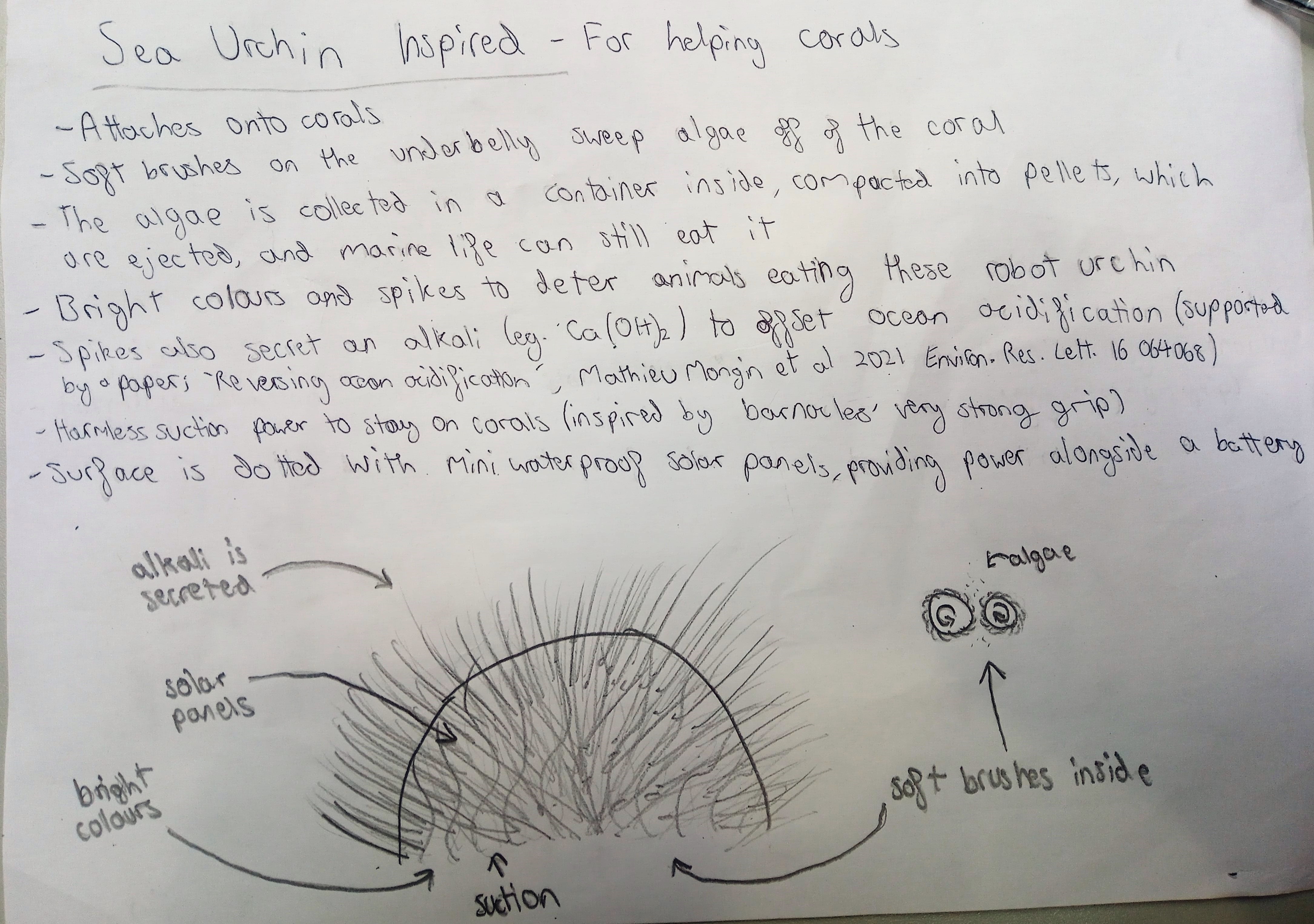}
      \caption{Sea Urchin-inspired submission by 'The Robotineers'. The pictured design is intended to protect corals by removing problematic algae and secreting alkaline substances.}
      \label{fig:urchin}
\end{figure*}

\begin{figure*}[h]
 	\centering
      \includegraphics[width=0.8\textwidth]{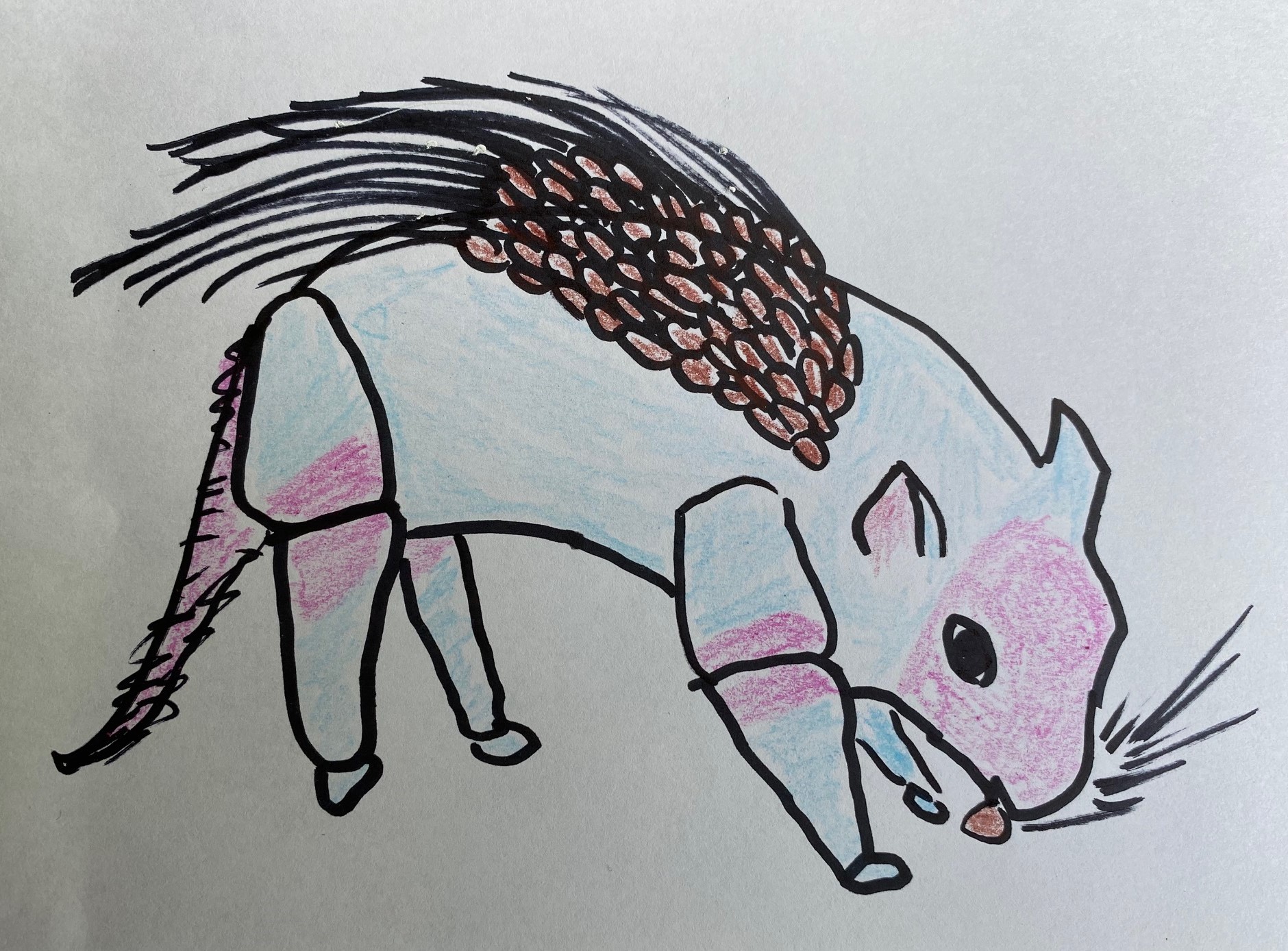}
      \caption{Contest Submission by Sue Klefstad. The pictured design is intended to employ the seed burying behaviours of squirrels as a way to plant Milkweeds.}
      \label{fig:milkweed}
\end{figure*}

\begin{figure*}[h]
 	\centering
      \includegraphics[width=0.8\textwidth]{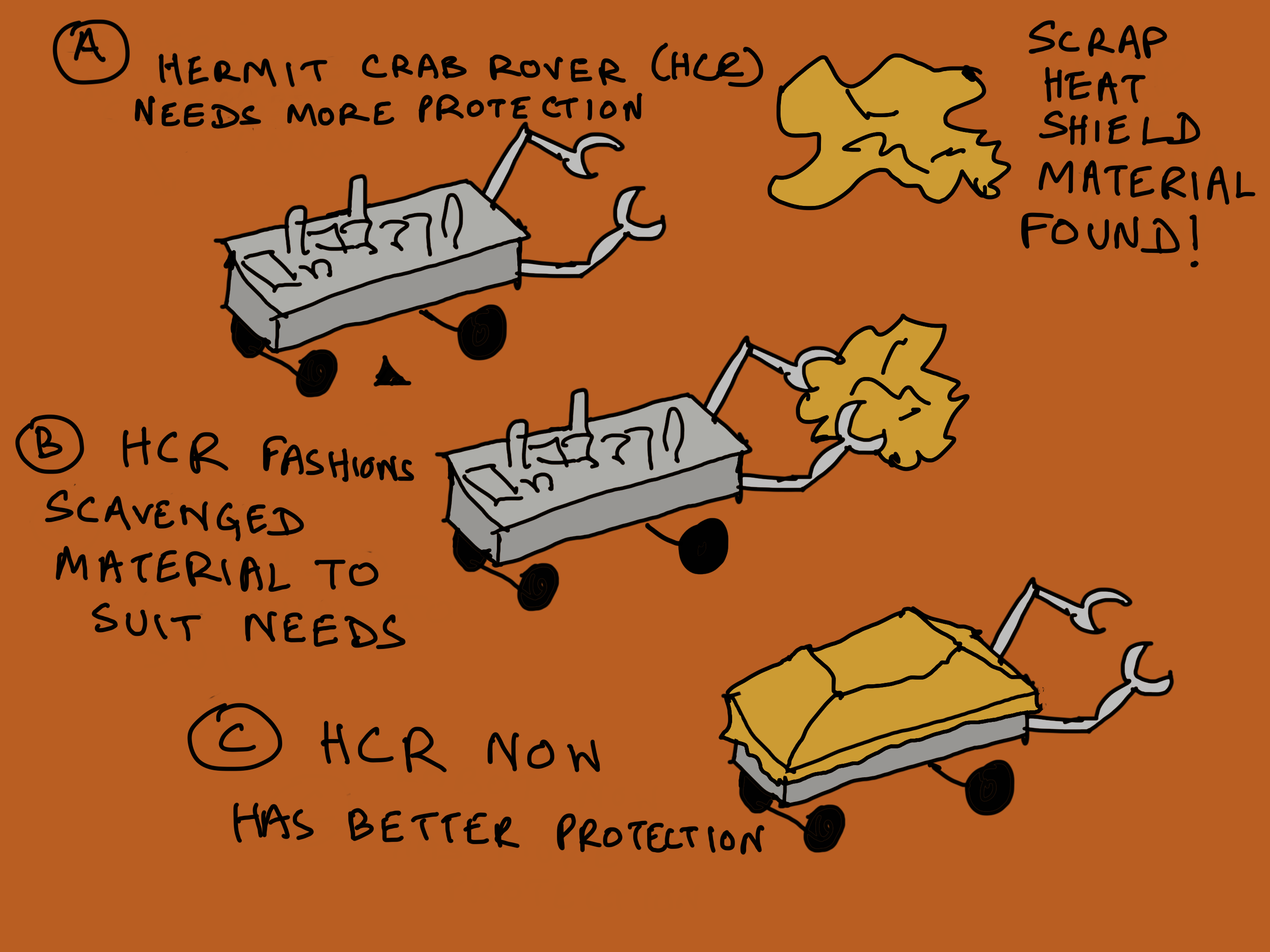}
      \caption{Contest Submission by 'The Yak Collective'. The pictured design is intended to scavenge material to form a protective shell around a mobile rover.}
      \label{fig:hermitcrab}
\end{figure*}

\begin{figure*}[h]
 	\centering
      \includegraphics[width=0.8\textwidth]{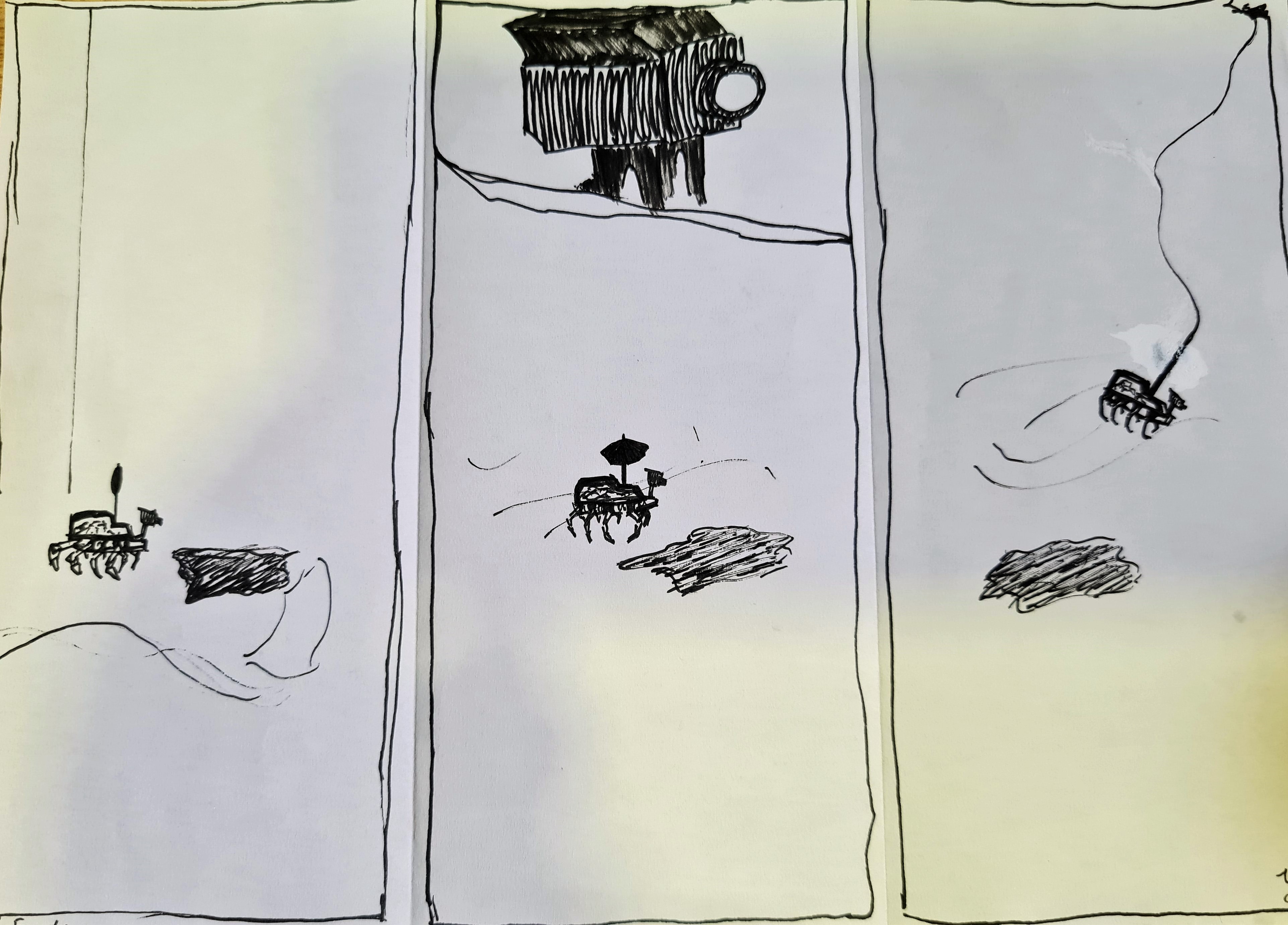}
      \caption{Contest Submission by Maier Fenster. A miniature robot emulates the ballooning behviour of spiders to move around its environment.}
      \label{fig:spiderpoppins}
\end{figure*}

\begin{figure*}[h]
 	\centering
      \includegraphics[width=0.8\textwidth]{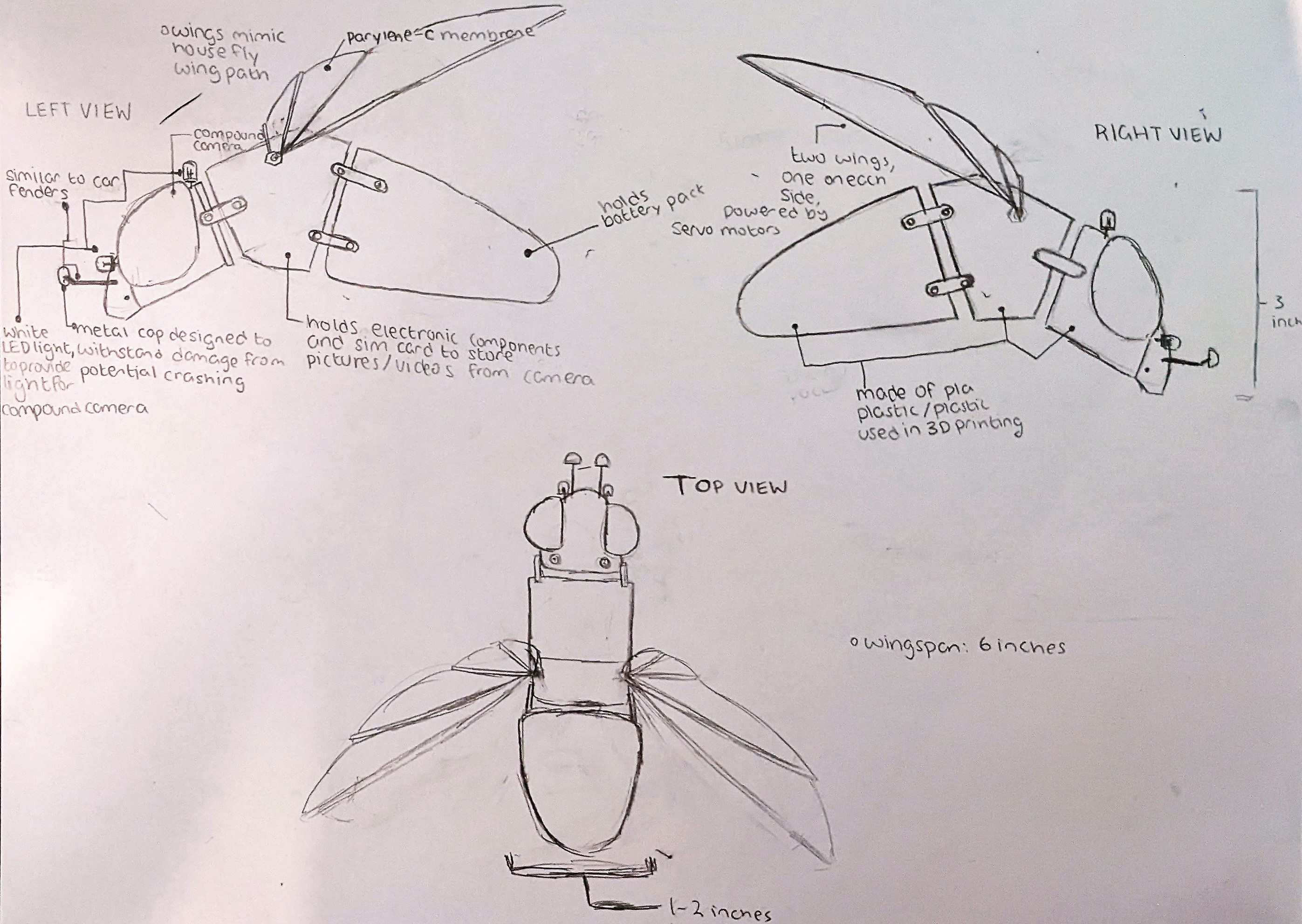}
      \caption{Contest Submission by Elizabeth Isaac. A miniature robot emulates the ballooning behaviour of spiders to move around its environment.}
      \label{fig:housefly}
\end{figure*}

\begin{figure*}[h]
 	\centering
      \includegraphics[width=1\textwidth]{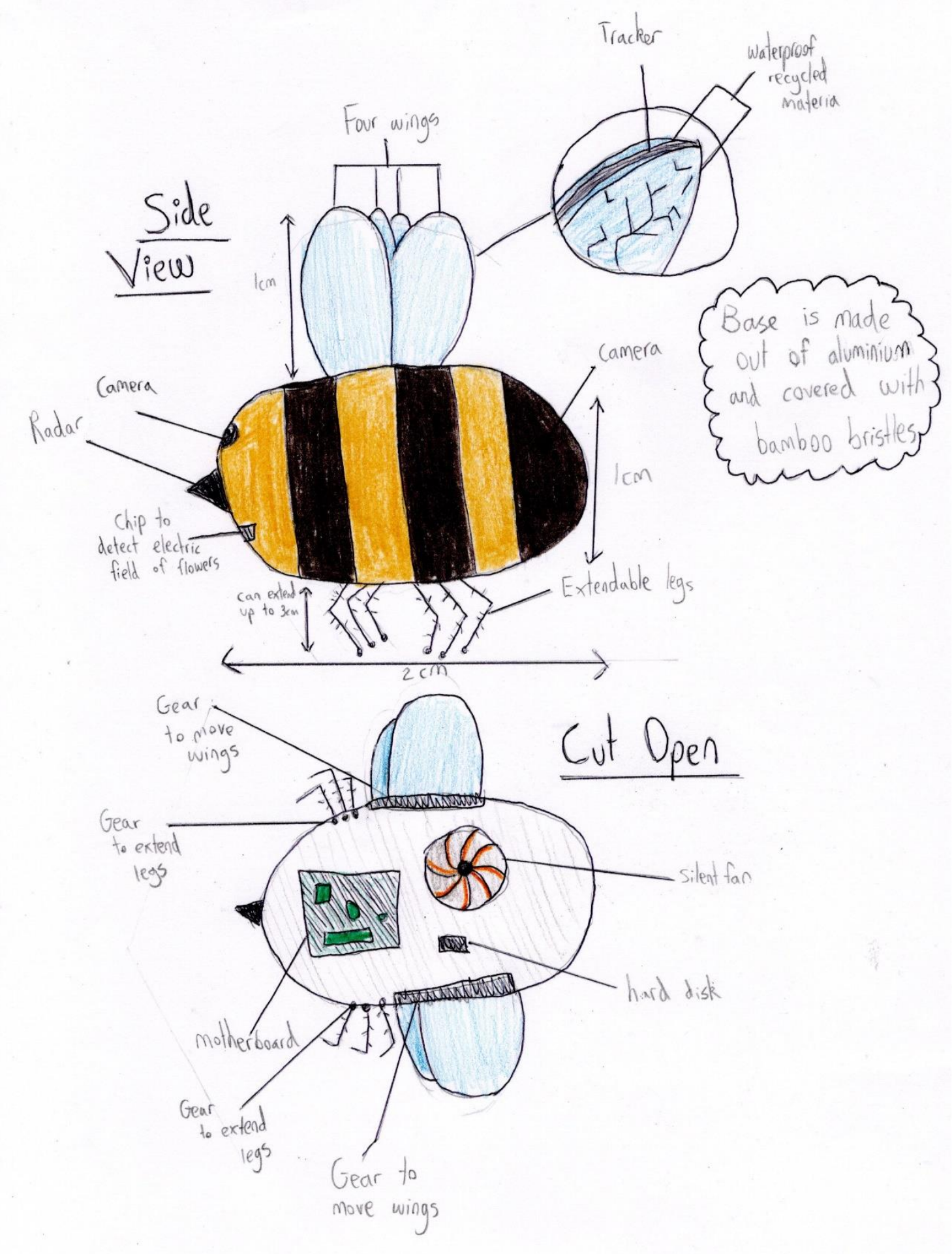}
      \caption{Contest Submission by Daniella Clifton. A robotic bumblebee pollinates plants}
      \label{fig:bumblebot}
\end{figure*}

\begin{figure*}[h]
 	\centering
      \includegraphics[width=1\textwidth]{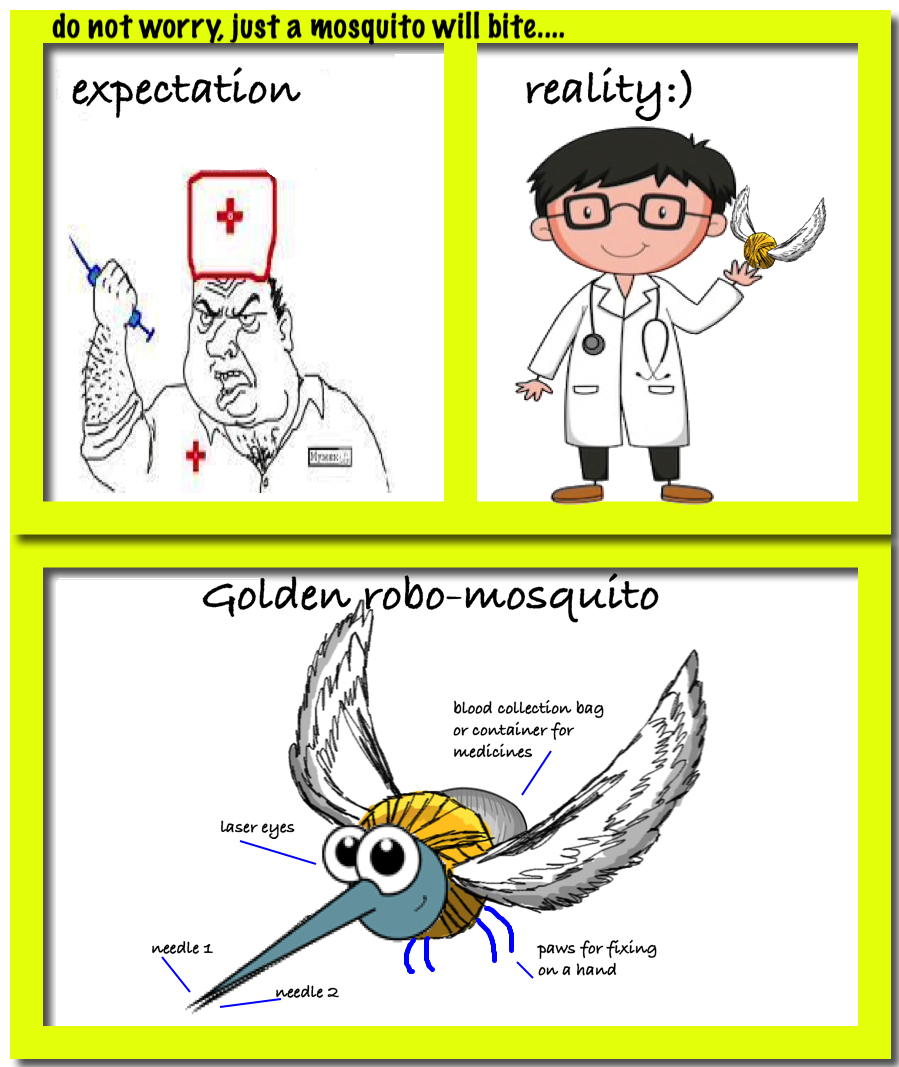}
      \caption{Contest Submission by Irina Putchenko. A robotic mosquito allows easier access to blood samples in medicine.}
      \label{fig:mosquito}
\end{figure*}

\clearpage

\section{Natural Robotics Contest Competition Advertisement} \label{sec:flyer}

\begin{figure*}[h]
 	\centering
      \includegraphics[width=0.9\textwidth]{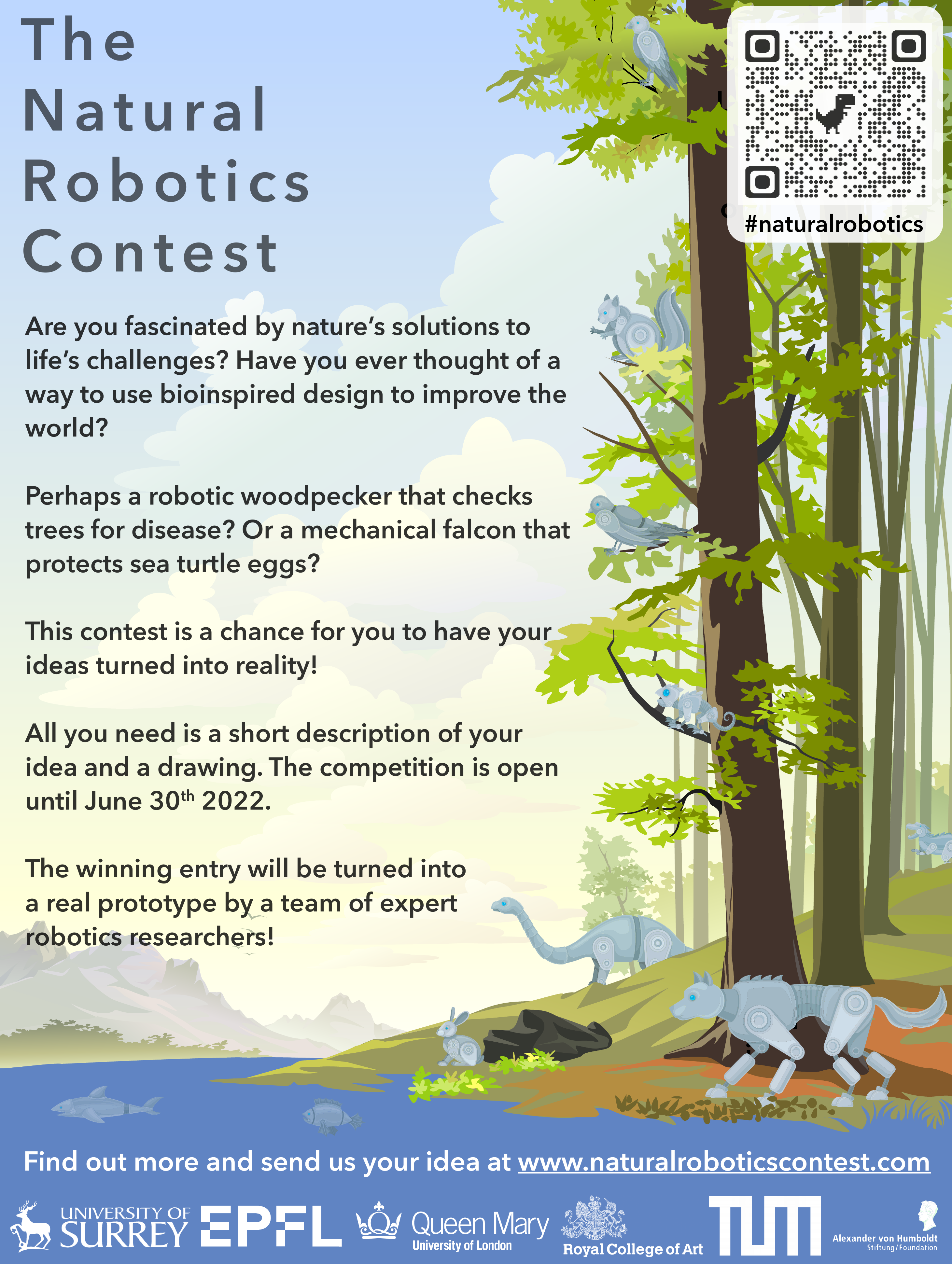}
      \caption{The contest was advertised across a variety of channels, including digital news media, university social media, Maker/Art forums, STEM outreach charities, and direct word-of-mouth. The flyer above was circulated widely for display on noticeboards and in email newsletters.}
      \label{fig:flyer}
\end{figure*}

\twocolumn
\section*{References}
\medskip

\bibliographystyle{vancouver}
\bibliography{Natrobib}

\end{document}